\documentclass[letterpaper, 10 pt, conference]{ieeeconf}  

\IEEEoverridecommandlockouts                              

\usepackage{graphics} 
\usepackage{subfig}
\usepackage{epsfig} 
\usepackage{mathptmx} 
\usepackage{times} 
\usepackage{amsmath} 
\usepackage{amssymb}  
\usepackage{color}
\usepackage{soul}
\usepackage[normalem]{ulem}
\usepackage{lipsum}
\usepackage[disable]{todonotes}

\newcommand{\comment}[1] {} 
\usepackage{amsmath} 
\usepackage{adjustbox}
\usepackage{multirow}
\usepackage{comment}
\usepackage[colorlinks,linkcolor=black]{hyperref}

\usepackage{algorithm}
\usepackage[noend]{algpseudocode}

\usepackage{array}

\usepackage{parselines}

\title{\LARGE \bf Human Driver Behavior Prediction based on UrbanFlow$^*$
}

\author{Zhiqian Qiao$^{1}$, Jing Zhao$^{2}$, Zachariah Tyree$^{3}$, Priyantha Mudalige$^{3}$, Jeff Schneider$^{4}$ and John M. Dolan$^{4}$ 
\thanks{*This work is supported by General Motors}
\thanks{$^{1}$Zhiqian Qiao is a Ph.D. student of Electrical and Computer Engineering, Carnegie Mellon University, 5000 Forbes Ave, Pittsburgh, USA
        {\tt\small zhiqianq@andrew.cmu.edu}}%
\thanks{$^{2}$ Mechanical Engineering, Carnegie Mellon University}
\thanks{$^{3}$ Research \& Development, General Motors}
\thanks{$^{4}$ Faculties of The Robotics Institute, Carnegie Mellon University}
}

\begin{document}

\maketitle
\thispagestyle{empty}
\pagestyle{empty}

\begin{abstract}

   How autonomous vehicles and human drivers share public transportation systems is an important problem, as fully automatic transportation environments are still a long way off. Understanding human drivers' behavior can be beneficial for autonomous vehicle decision making and planning, especially when the autonomous vehicle is surrounded by human drivers who have various driving behaviors and patterns of interaction with other vehicles. In this paper, we propose an LSTM-based trajectory prediction method for human drivers which can help the autonomous vehicle make better decisions, especially in urban intersection scenarios. Meanwhile, in order to collect human drivers' driving behavior data in the urban scenario, we describe a system called UrbanFlow which includes the whole procedure from raw bird's-eye view data collection via drone to the final processed trajectories. The system is mainly intended for urban scenarios but can be extended to be used for any traffic scenarios.

\end{abstract}

\section{INTRODUCTION}

A major challenge in recent work on autonomous vehicles is making proper decisions about how to deal with interactions with human-driven vehicles. However, interactions among human drivers are hard to model via equations directly. To address this problem, learning-based methods for characterizing human-driver behavior become good choices and make it easier to simulate a human driver's behavior in simulations such as CARLA \cite{carla}, VTD \cite{vtd}, etc. However, such methods require a large amount of driving data in order to learn human drivers' diverse behavior. For a long time, NGSIM \cite{ngsim} was the only public trajectory-based dataset from which human driver behavior could be extracted. In 2018, highD \cite{highd} became available, but it only includes highway scenarios. Moreover, how to extract and classify the human driver behavior without manually labeling a large amount of data for ground-truth is another time-consuming challenge when dealing with raw human driver data.

\begin{figure}[!t]
\centering
\includegraphics[width=\columnwidth]{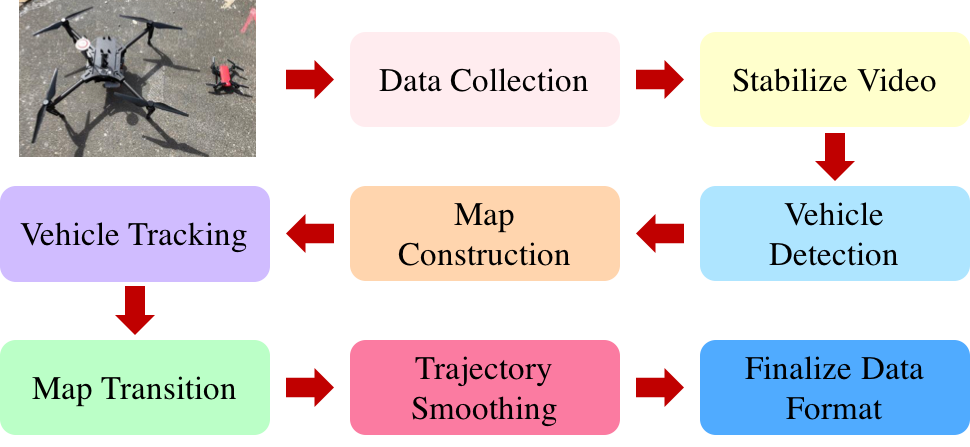}
\caption{The UrbanFlow dataset processing pipeline. The pipeline includes the drone data collection and process flow from raw video data to the final trajectory data.}
\vspace{-0.5cm}
\label{fig:pipeline}
\end{figure}

The current state of the art in acquiring and using such data faces several problems. First, some published work relies on privately collected datasets, the inaccessibility of which makes them impossible to use as benchmarks for comparisons between various algorithms. Second, some datasets are collected by autonomous vehicles from the perspective of the ego vehicle. Although this perspective is ultimately the one available to an autonomous vehicle, it is difficult for it to provide full sequences showing the social behavior of surrounding vehicles. To derive models for such behavior, bird's-eye view datasets are useful. In response to these problems, this paper constructs a method for benchmarking human driver behavior based on a bird's-eye-view data collection system via drone. Figure \ref{fig:pipeline} shows the pipeline of the data processing procedure.

On the other side, based on the dataset, predicting the other vehicles' intentions or trajectories is an essential procedure for behavior planning of autonomous vehicles during the decision making or trajectory planning procedures. During the application of motion planning practice, accurately predicting human drivers' behavior can help the ego car to have better decision making. In Pittsburgh, most traffic lights control \textit{Going Straight} (\textit{GS}), \textit{Turning Left} (\textit{TL}) and \textit{Turning Right} (\textit{TR}) with one light with the result that at the urban intersection, many interactions occur between vehicles approaching from opposite directions with intention pair of \textit{GS} and \textit{TL} or \textit{TL} and \textit{TR}. Under these situations, 'who will go first' between the two interacting vehicles is a key problem even for human drivers.

The main contributions of this work are:

\begin{itemize}

\item A drone-based data collection and processing system to analyze bird's-eye view trajectory data of human drivers.

\item An algorithm which can predict the interacting human drivers' intentions as well as trajectories based on the historical trajectories occupying a given period of time when approaching an an urban intersection.

\end{itemize}

\section{Related Work}

This section introduces previous work related to this paper, which can be categorized as follows: 1) papers that address algorithm which is part of the traffic data collection procedures; 2) papers that propose intention and trajectories predictions of human drivers.

\subsection{Data Extraction}
With the current popularity of autonomous driving, various datasets are available for researchers to develop and test their algorithms. These datasets can be categorized into two classes. The first one is traffic-flow-based datasets, which focus on a particular scene and simultaneously capture all the vehicles within it. This type of dataset uses a bird's-eye view to observe vehicle trajectories within the scene. The NGSIM dataset \cite{ngsim} is the best-known such dataset and includes highway and urban scenarios. Last year, RWTH Aachen University released the highD dataset \cite{highd}, which used advanced computer vision technology to improve the data collection mechanism based on the NGSIM dataset. Another kind of dataset is based on the sensors mounted on the ego-vehicle and data collected while driving the ego car over a given route. Most such datasets create various vision-based benchmarks for further study. The KITTI dataset \cite{kitti} offers a vision benchmark for different autonomous vehicle-related tasks. The Oxford RobotCar dataset \cite{oxford} collected 20 million images from 6 cameras mounted on the vehicle, along with LIDAR, GPS, and INS ground truth. Recently, UC Berkeley released the BDD100k \cite{bdd} which includes diverse driving videos collected from a camera mounted on the vehicle with scalable annotation tooling.

In the current work, in order to gain a comprehensive view of the traffic situation, we use a bird's-eye-view method to collect traffic-flow-based datasets via drone. The portable end-to-end system allows researchers to collect their own data from any site of interest, unlike the NGSIM system, which depended on the installation of a fixed camera. While our data collection method is similar to that used for the highD dataset, our method focuses primarily on urban intersections, which are more challenging than the highway scenarios that the highD dataset focuses on. Compared with highD, this work extends to urban scenario.

\subsection{Prediction}

Liebner \cite{driverintent} proposed an explicit model to extract characteristic desired velocity profiles from real-world data that allow the Intelligent Drive Model (IDM) to account for turn-related deceleration to represent both car-following and turning behavior. Derek et al. \cite{generalizableintention} used LSTMs to classify vehicle maneuvers at intersections. They predicted whether a driver will turn left, turn right, or continue straight up to 150m with consistent accuracy before reaching the intersection using LSTM, with the mean cross-validated prediction accuracy averaging over 85\% for both three- and four-way intersections. There are other works on predicting complete trajectories using Hidden Markov Models, Gaussian Processes, Dynamic Bayesian Networks, Support Vector Machines, and inverse reinforcement learning. 

Compared with \cite{driverintent}, besides velocity profile, multiple factors are added in our models, such as yaw variation, target motion features, etc., which contain information on environmental changes for the ego vehicle. 
The work concentrates on the interaction of ego and target car pairs by studying the related interaction with each other. Meanwhile, we introduce the idea of direction intention prediction and use the result to determine a more detailed trajectory prediction. The main difficulties we tackled in the work is that the human driver's intentions and trajectories of the vehicle are much more variable when approaching an urban intersection with heavy traffic flow than in highway situations.

\section{Preliminaries}

In this section, the preliminary background of the problem is described. The fundamental algorithms which are used for video stabilization, object detection and tracking are included here.

\subsection{Enhanced Correlation Coefficient}

The two main challenges for video stabilization are the robustness and the speed of the alignment. Feature-based alignment is fast and is able to align images with large displacement. However, its robustness is susceptible to the quality and distribution of the feature points detected. On the other hand, the image alignment algorithm like Enhanced Correlation Coefficient (ECC) \cite{ecc} can be used to process every pair of two consecutive frames in the video. But each alignment iteration needs all the pixels to be searched, which is computationally expensive. Moreover, ECC also fails to align frames without a good initial guess at the homography matrix when the two frames have a low similarity. As a result, a combination of feature-matching-based and homography-based methods is used to reap the advantages of both.

Feature-based alignment involves detecting key-points, finding key-point correspondences, and computing image transformation using the Random sample consensus (RANSAC) \cite{RANSAC} algorithm. The standard ECC alignment uses normalized intensity with zero mean so that the similarity measurement is invariant to contrast and brightness change \cite{ecc}. Each frame was warped first using the homography calculated from the feature-based method for a rough alignment, then warped with homography calculated from ECC alignment.

\subsection{Retinanet}

In the proposed pipeline, RetinaNet \cite{retinanet} is used for detecting vehicles in the images. RetinaNet has a backbone network which is responsible for computing the convolution feature map over an entire input image. A class sub-net is responsible for predicting the class of object. In our case, there is only one class, which is 'car'. A box regression sub-net predicts the location and size of each vehicle. ResNet-50 was used as the backbone for the forward pass of the FPN architecture since the residual learning framework promotes easier convergence.

\subsection{Kalman Filter}

A Kalman filter \cite{kf} is used for tracking and trajectory smoothing. Based on the car's dynamic model, characteristics of the system noise and measurement noise, the measurement variables are used as the input signal, and the estimation variables that we need to know are the output of the filter. The whole filtering process is composed of a prediction equation and an update equation as follows:
\begin{equation}
    \begin{split}
        X(n) & = F X(n-1)  + V_q(n-1)\\
        Y(n) & = H X(n) + V_p(n)\\
    \end{split}
\end{equation}
\noindent
where $X (n)$ and $Y (n)$ are the estimated state variable and measurement variable at frame $n$, respectively. 

\section{Methodology}

In this section we propose UrbanFlow as a procedure to deal with the collected bird's-eye view data. Then, based on UrbanFlow, we propose a method for predicting the human driver's intention as well as trajectories.

\subsection{UrbanFlow}

\subsubsection{Video Stabilization}
\label{sectionstable}

\begin{figure}[!t]
      \centering
      \includegraphics[width=\columnwidth]{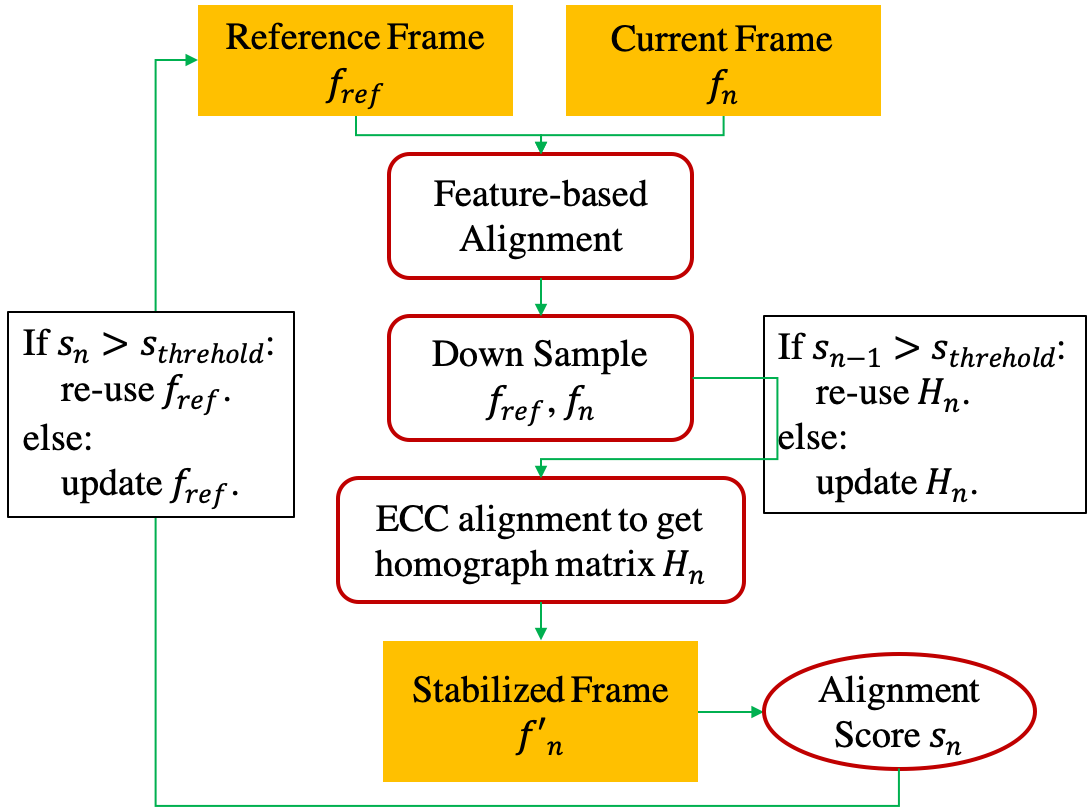}
      \caption{Optimized stabilization method flow.}
      \label{stable}
      \vspace{-0.5cm}
   \end{figure}

In this paper, we propose several steps for the video stabilization in order to deal with the displacement of the drone during the data collection process. Figure \ref{stable} visualizes the flow of the stabilization method. For each frame $f_n$ at time step $n$, the algorithm chooses a reference frame $f_{ref}$ according to the alignment evaluation score gotten from the result of the last time step and corresponding homograph matrix in order to get the stabilized frame. Firstly, a re-alignment is performed when the result of the ECC alignment score is lower than a threshold. ECC takes a lot of time to converge and is not adaptive to align the current frame with a reference frame when their similarity is lower than a threshold. Secondly, since the alignment is time-consuming, it is only performed when a reference frame needs to be re-chosen. The homography matrix is re-used for the following frames until a new reference frame is chosen when the evaluation score drops to the threshold. Then, the homography matrix calculated from ECC alignment during the previous step is used for initializing the guess for ECC in the next step to speed up the convergence. Lastly, images are down-sampled \cite{ECC_Downsample} so that ECC uses fewer pixels during the calculation. 


\subsubsection{Object Detection}

The training dataset contains all the bounding boxes and their corresponding labels for each image. The input images are re-sized to ensure that the size of detection objects is greater than 32-by-32 pixels as well as not too large to fit for the GPU computational capability. Images are masked to crop out the roads in order to make detection easier. RetinaNet was fine-tuned using pre-trained weights from the COCO dataset \cite{coco}.

\subsubsection{Map Construction and Coordinate Transition}

  \begin{figure}[!t]
      \centering
      \includegraphics[width=\columnwidth]{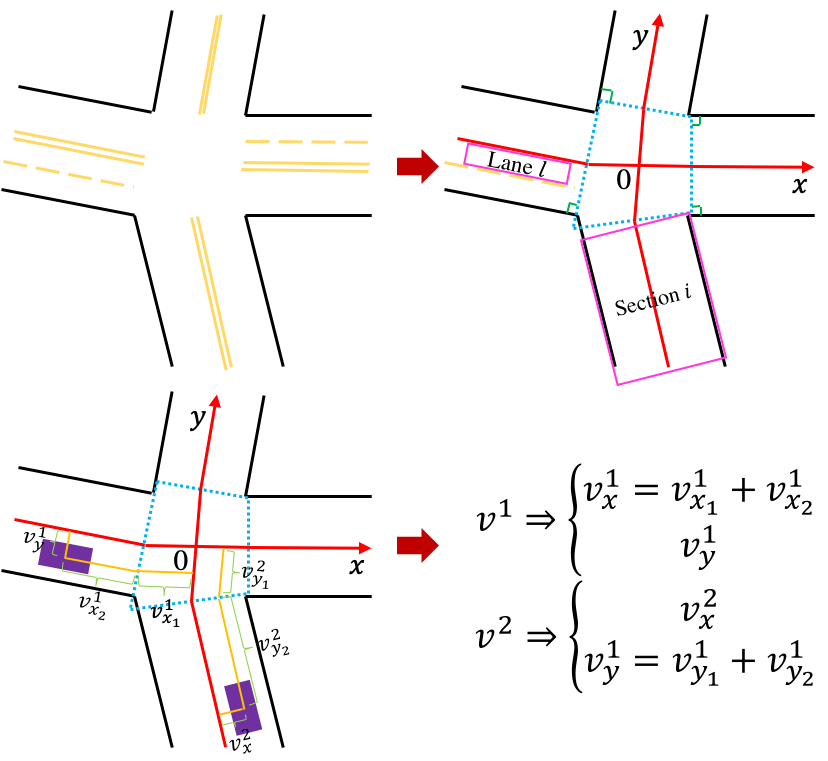}
      \caption{Transition from original image-based coordinate to road-based Coordinate}
      \label{coor}
      \vspace{-0.5cm}
   \end{figure}
   
The first step in the creation of the map is to crop the area of interest, which in this case is the roads. To attack this problem, we took advantage of the image segmentation network "U-net", described in Ronneberger et al. \cite{ronneberger2015u}, with just a few adjustments based on the work of Iglovikov et al. \cite{iglovikov2018ternausnet}. We preserve the decoder section of the network because by adding a large number of feature channels, it allows the network to propagate context information to the higher-resolution layers. The important change was in the encoder section, where it was replaced by the down-sampling elements of the VGG16 architecture in order to take advantage of the pre-trained weights in ImageNet \cite{imagenet}, due to the limitation of the quantity of the collected data.

After detecting the road and applying a color filter to detect the lane markings on the road, the work transforms all the detected vehicle positions from the original image-based coordinates to the road-based coordinates. Figure \ref{coor} shows the method to generate the road-based coordinate based on a random road geometry which may occur in the real world. The method firstly chooses an origin and then proceeds to obtain the $x$ axis and $y$ axis along the lane markings which separate the opposite directions of moving vehicles. For the given vehicles $v^1$ and $v^2$, the figure lists two examples of how to extract the road-based positions. Finally, it is able to represent the vehicles' information, which contains the following items:

\begin{itemize}
\item Local $x$ and $y$ based on the road-based coordinates
\item Vehicle length and width
\item Section ID $i$
\item Lane ID $l$
\end{itemize}

\subsubsection{Vehicle Tracking and Trajectory Smoothing}

After the positions of vehicles have been transformed into the local (road) coordinates, we apply the tracking algorithm to track each car. Meanwhile, we smooth each vehicle's trajectory.

In the system, we use vehicle position as the state variable. $F$ is the state transition matrix and $H$ is the measurement matrix. $V_q(n)$ and $V_p(n)$ represent the system noise and measurement noise, respectively.

\subsection{Driving Behavior Prediction}

\subsubsection{Intention Prediction}

\begin{figure}[!t]
    \centering
    \includegraphics[width=\columnwidth]{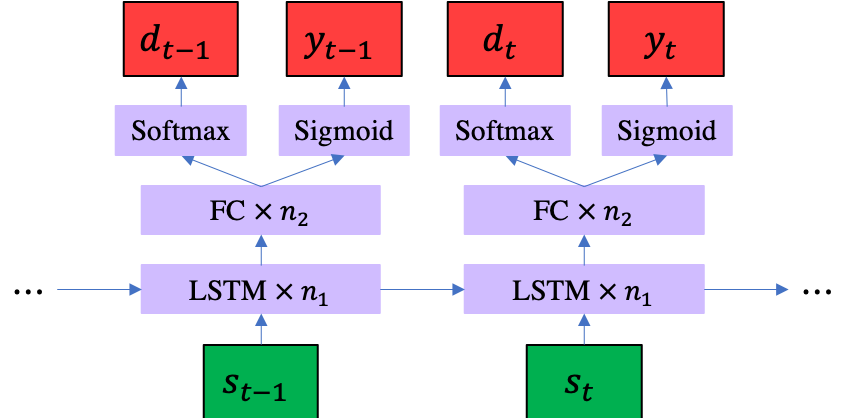}
    \caption{Intention Prediction Network Structure}
    \label{fig:networki}
 \end{figure}
 
 \begin{figure}[!t]
    \centering
    \includegraphics[width=\columnwidth]{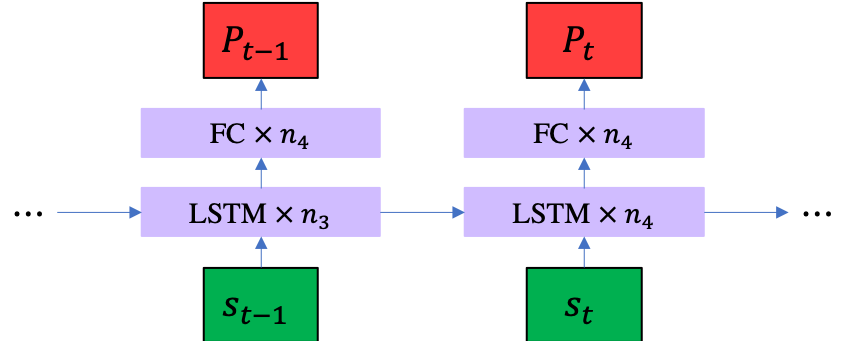}
    \caption{Trajectories Prediction Network Structure}
    \label{fig:networkt}
    \vspace{-0.5cm}
 \end{figure}

For the driving behavior task, we construct the network with an LSTM layer which gets the Direction Intention $d$ and Yield Intention $y$ (see Figure \ref{fig:networki}). The direction intentions include \textit{Going Straight} (\textit{GS}), \textit{Turing Left} (\textit{TL}) and \textit{Turning Right} (\textit{TR}). The yield intention indicates the prediction of which car will go through the potential crash point first. For the interacting driver pairs with intentions of \textit{GS} and \textit{TL} or \textit{TL} and \textit{TR}, the input states include the positions, velocities, heading angles and relative distance to the intersection center of both cars of each pair. During the interaction procedure, the yield motion also changes based on the counterpart's behavior. This will contribute as a key factor to the next-step motion planning module and help to generate a safer and feasible trajectory.

\subsubsection{Trajectories Prediction}

 \begin{figure}[!t]
    \centering
    \includegraphics[width=0.7\columnwidth]{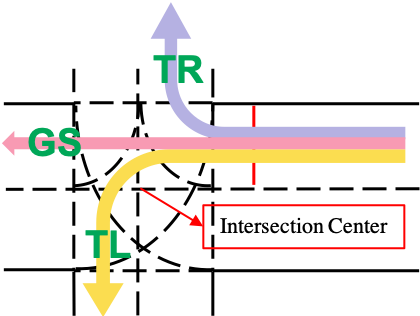}
    \caption{Reference trajectories according to the direction intention. Follow the center of the lanes.}
    \label{fig:reference}
 \end{figure}

Based on the results of direction and yield predictions, a more detailed trajectory prediction procedure includes more information on the future trajectories. In Figure \ref{fig:networkt}, $P_t$ includes information on velocities and positions of the target car. A reference trajectory is first selected according to the intention prediction results. According to the reference trajectories (see Figure \ref{fig:reference}) with intersection geometry information, the velocities, heading angles and relative distance to the intersection center of both cars, the network can predict the future trajectories.

\section{Experiment}

In this section, we show the results for methods corresponding to different data processing procedures.

\subsection{UrbanFlow}

\begin{table}[!t]
\caption{Comparison between different stabilization methods. DS means down sampled. }
\label{ecc}
\begin{center}
\begin{tabular}{c |c| c}
\hline
Method & Processing Time ($s/frame$) & SSIM\\
\hline
ORB $+$ ECC w/o DS & 1.7609 & 0.8032\\
ORB $+$ ECC, $\frac{1}{2}$ DS & 0.6724 & 0.7759\\
ORB $+$ ECC, $\frac{1}{4}$ DS & 0.4779 & 0.7324\\
ORB $+$ ECC, $\frac{1}{8}$ DS & 0.4071 & 0.7160\\
SURF $+$ ECC w/o DS & 0.6599 & 0.81896\\
SURF $+$ ECC, $\frac{1}{2}$ DS & 0.6627 & 0.7758\\
SURF $+$ ECC, $\frac{1}{4}$ DS & 0.4960 & 0.7336\\
SURF $+$ ECC, $\frac{1}{8}$ DS & 0.3750 & 0.7166\\
ECC, $\frac{1}{2}$ DS & 0.9.3450 & 0.9404\\
ORB & 3.4665 & 0.8278\\
SIFT & 13.575 & 0.8370\\
SURF & 2.1060 & 0.8390\\
\hline
\end{tabular}
\end{center}
\vspace{-0.75cm}
\end{table}

\subsubsection{Video Stabilization}
In the previous section, we have introduced the combination of the feature-matching-based and homography-based alignment methods. The result compares different combinations of feature-matching-based and homography-based video stabilization algorithms with various down-sampling ratios. Table \ref{ecc} shows the results of different choices of algorithms and the corresponding structural similarity (SSIM) score which is used to calculate the similarity between any two images. The higher SSIM score indicates a better stabilization result.

\begin{figure}[!t]
\centering
\includegraphics[width=\columnwidth]{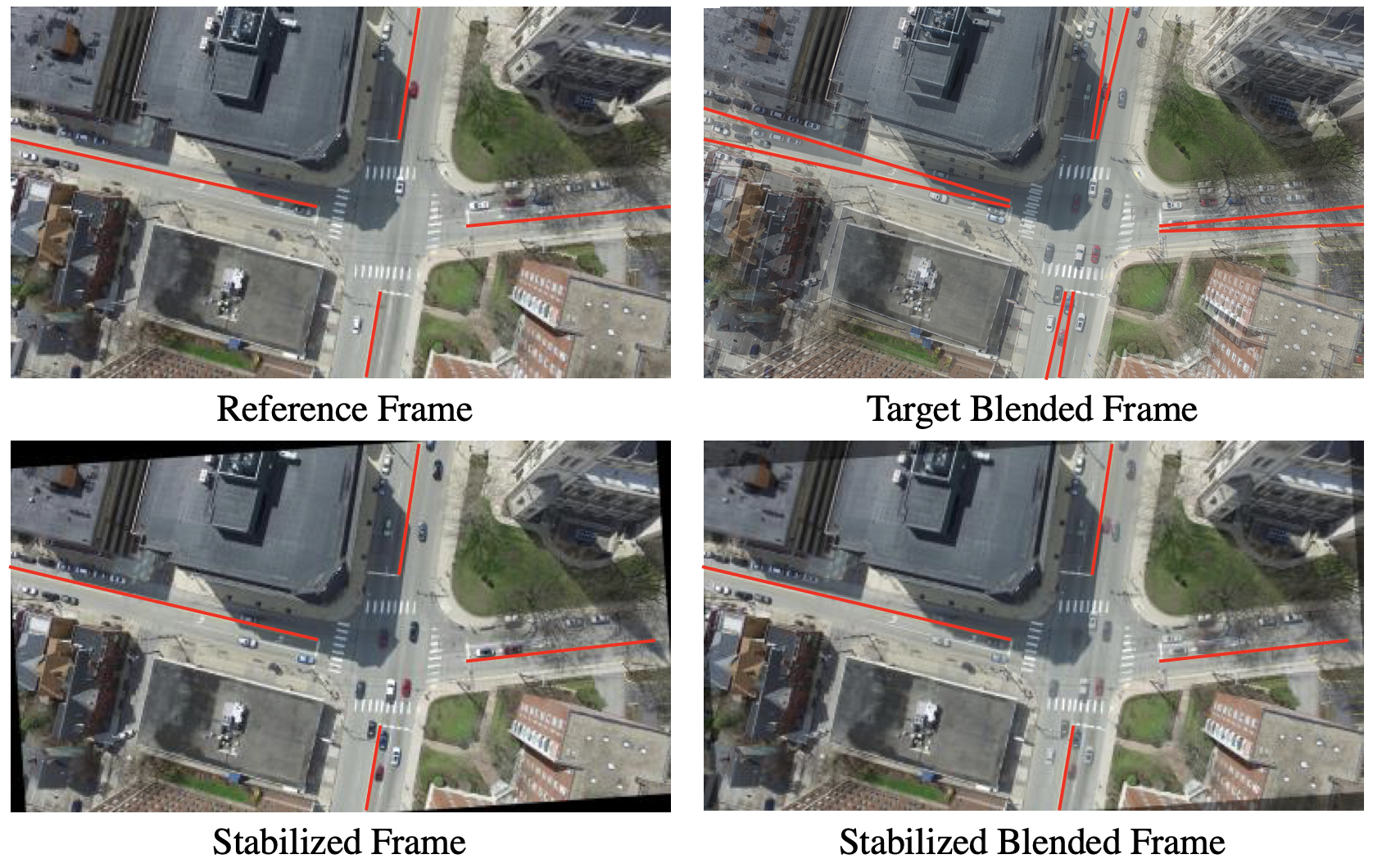}
\caption{Two blended frames before stabilization and after stabilization}
\label{img:stable}
\vspace{-0.5cm}
\end{figure}

We finally chose the Speeded Up Robust Features (SURF) detector combined with ECC and $\frac{1}{8}$ down-sampling to get a relatively good tradeoff between stabilization and computational efficiency. We visualize images with and without stabilization in Figure \ref{img:stable} with four sub-figures. The Reference Frame shows the anchor frame for the stabilization. Ideally, the roads can be perfectly aligned in the Reference Frame and Target Frame. Before stabilization, the Reference Frame and the Target Frame are blended, which is shown as the Target Blended Frame. It is obvious that the two frames have a big misalignment. After the stabilization of the frame, the result is shown as the Stabilized Frame and then the new blended result is shown as the Stabilized Blended Frame.

\subsubsection{Vehicle Detection}

\begin{figure}[!t]
\centering
\includegraphics[width=\columnwidth]{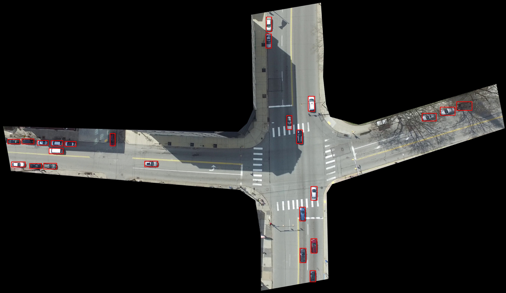}
\caption{Vehicle detection results of Retinanet algorithm for a selected frame.}
\label{img:detection}
\end{figure} 

\begin{table}[!t]
\caption{Detection Result}
\label{table:retinanet}
\begin{center}
\begin{tabular}{c |c| c}
\hline
P / T  & Car  (Train / Test)& No Car (Train / Test) \\
\hline
Car & 387 / 2369 & 1 / 0 \\
No Car  & 3 / 72  & 0 / 0\\
$\frac{GT \cap PR }{GT \cup PR }$ & \multicolumn{2}{c}{0.95 / 0.994  }\\
$\frac{GT \cap PR }{GT }$ & \multicolumn{2}{c}{0.97 / 0.995 }\\
$\frac{GT \cap PR }{ PR }$ & \multicolumn{2}{c}{0.97 / 0.996 }\\
\hline
\end{tabular}
\end{center}
\vspace{-0.5cm}
\end{table}

By using Retinanet to do vehicle detection, we trained a good model to detect vehicles from a bird's-eye view. For testing, only 97 out of 2322 vehicles are not detected, giving an accuracy of $96\%$, and the average intersection over union is $92\%$. This accuracy is high since the vehicles in the test cases are similar to the ones during training. False positives were removed using non-maximum suppression and thresholding the confidence score for a prediction. If vehicles such as a bus appear in testing but had never appeared in training, these vehicles will not be detected, since they are too different from what the network has learned both in size and color. The images with incorrect detection were relabeled to fine-tune the network.

For each frame, RetinaNet is applied to detect vehicles. Figure \ref{img:detection} visualizes one of the testing images after applying the vehicle detection method. The original image-based positions of all the red bounding boxes detected as vehicles are saved. Table \ref{table:retinanet} shows the quantitative results of training and testing. $GT$ is the abbreviation for the area of Ground Truth and $PR$ is the abbreviation for the area of Predicted Results.

\subsubsection{Trajectory Smoothing}

\begin{figure}[!t]
\centering
\includegraphics[width=0.7\columnwidth]{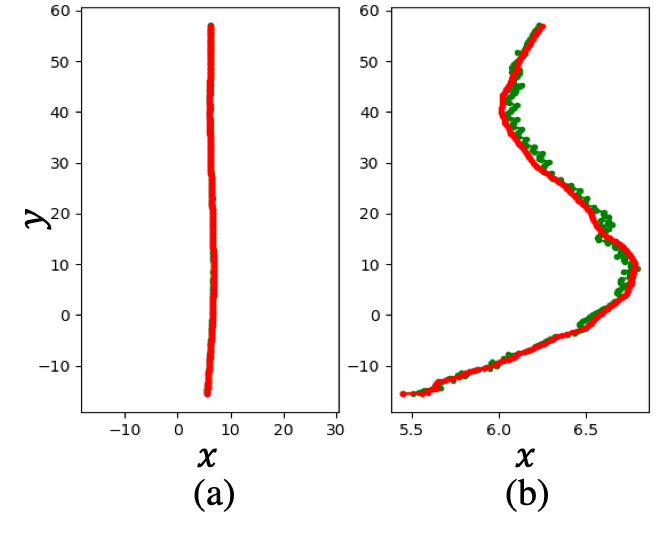}
\caption{Comparison between vehicle trajectories with and without smoothing.}
\label{img:smooth}
\vspace{-0.5cm}
\end{figure}

Most existing vehicle trajectory datasets, such as NGSIM \cite{ngsim}, only provide raw trajectories, which are noisy and therefore hard to use directly due to the jerky trajectories. Figure \ref{img:smooth} visualizes the results of the trajectories for one of the vehicles with and without smoothing. The Figure \ref{img:smooth}(a) shows the result with equal scaling of the $x$ and $y$ axes. It is hard to find the difference between the trajectories with (RED) and without (GREEN) smoothing. However, when the $x$ axis is enlarged in figure \ref{img:smooth}(b), the trajectory without smoothing (GREEN) is much jerkier than the one with smoothing (RED).

Finally, the video\footnote{\url{https://youtu.be/oTPgLUdN_cU}} includes all the dynamic results proposed in the pipeline.

  \begin{figure}[!t]
    \centering
    \includegraphics[width=0.8\columnwidth]{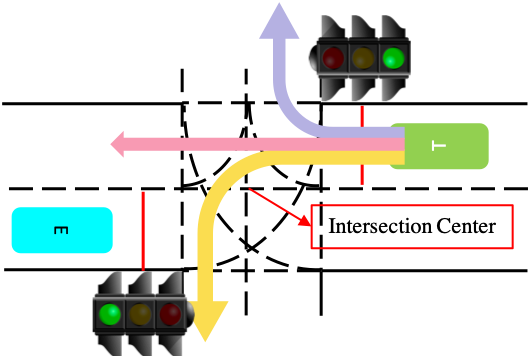}
    \caption{Scenario of interacting vehicles pair.}
    \label{fig:sce}
    \vspace{-0.5cm}
 \end{figure}

\subsection {Prediction}
 \begin{figure*}[!t]
    \centering
    \includegraphics[width=\textwidth]{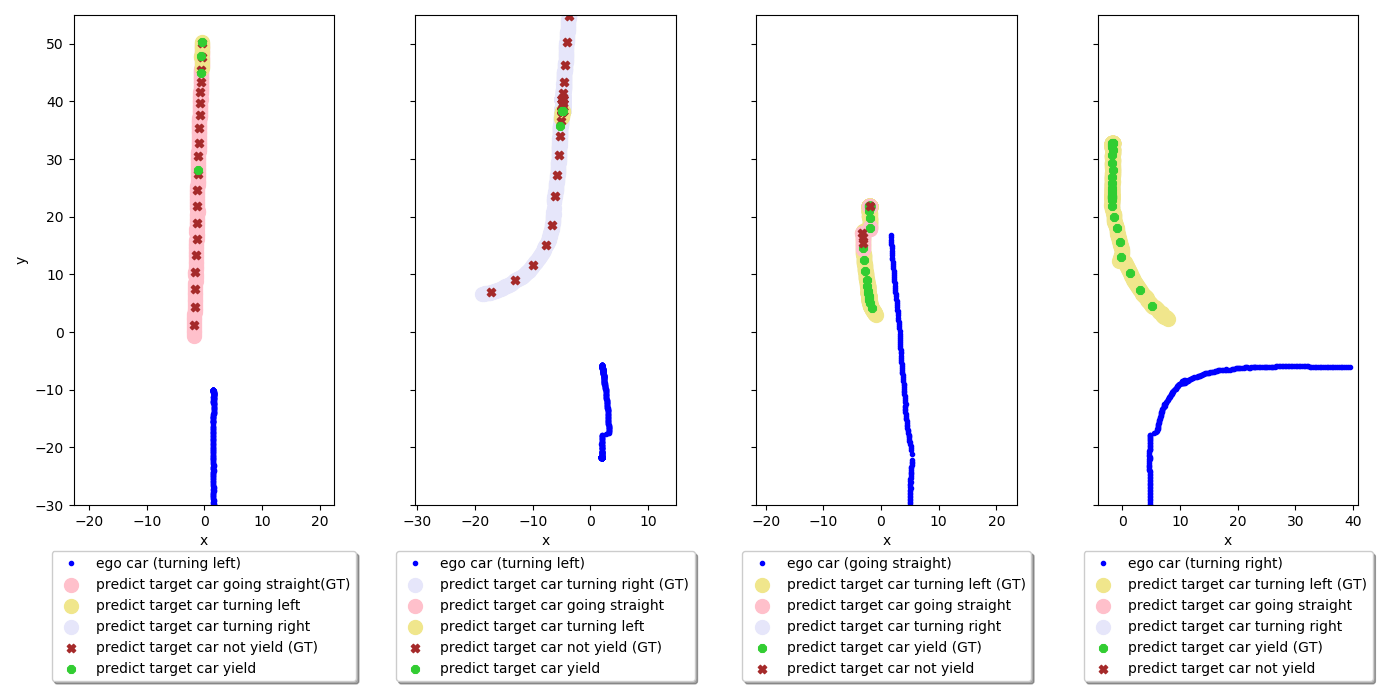}
    \caption{The direction and yield prediction result of selected interacting pairs. GT means that the corresponding intention is the ground truth of the selected pair. Direction intention of ego cars is noted in parentheses in the ego car legend.}
    \label{fig:intention_result}
 \end{figure*}
 
\subsubsection{Scenario}

We tested the algorithm based on the UrbanFlow dataset. We selected pairs of interacting vehicles with the driving directions of \textit{GS} and \textit{TL} or \textit{TL} and \textit{TR} from the dataset. Figure \ref{fig:sce} shows a pair of two interacting vehicles. The blue rectangle with $E$ is the ego car and the green rectangle with $T$ is the target vehicle. The input state includes velocities, heading angles and relative distances to the intersection center of both ego and target cars. 

\subsubsection{Intention Prediction}

 \begin{figure}[!t]
    \centering
    \includegraphics[width=\columnwidth]{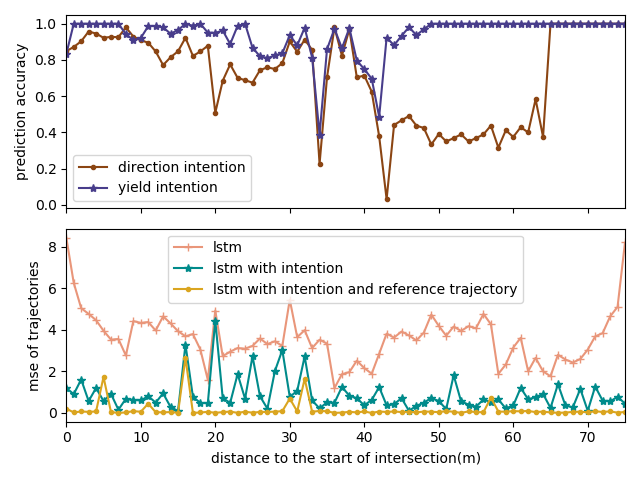}
    \caption{Direction and yield intention as well as MSE of trajectories prediction for the target vehicle.}
    \label{fig:intention_result_2}
    \vspace{-0.5cm}
 \end{figure}

According to the described state, the intention network predicts the direction intention as well as the yield intention. Figure \ref{fig:intention_result} visualizes the results of direction and yield intentions. According to the coordinate transitions, all the ego cars approach the intersection (intersection center is coordinate $(0, 0)$) from the bottom. Different colors with marker $\bullet$ show the results of direction predictions when the target vehicle reaching that position and the other colors with marker $\textbf{X}$ present the yield intention prediction results. Figure \ref{fig:intention_result_2} shows the prediction accuracy with respect to the distance to the start of intersection for the target vehicle.

\subsubsection{Trajectory Prediction}

We compared the mean squared error (MSE) results between the trajectory prediction with and without intention results and reference trajectories. Table \ref{table:mse} presents the average MSE for different methods and Figure \ref{fig:intention_result_2} shows the MSE with respect to the distance to the start of intersection for the target vehicle. When pass the start position of the intersection, the trajectories become diverse due to various direction intentions, as a result, the MSE of trajectories prediction increase.

\begin{table}[!t]
\caption{MSE of trajectory predictions.}
\label{table:mse}
\begin{center}
\begin{tabular}{c |c}
\hline
Method & Average MSE (m) \\
\hline
LSTM & 3.71\\
LSTM w/ intention & 0.89 \\
LSTM w/ intention and reference trajectory & 0.18\\
\hline
\end{tabular}
\end{center}
\vspace{-0.5cm}
\end{table}

\section{CONCLUSIONS}

In this paper, we propose a pipeline called UrbanFlow which is used to deal with traffic data collected by drones in urban environments. The raw data are processed through video stabilization, vehicle detection, map construction and coordinate transformation, vehicle tracking, and trajectory smoothing. Moreover, the paper proposes a method for driving behavior clustering and tests it on the UrbanFlow dataset. The following work for improving the dataset will focus on increasing the quantity of the dataset. More types of urban scenarios like T-intersections, stop-sign intersections and yield intersections will be included in the dataset.



\addtolength{\textheight}{-14cm}   





\bibliographystyle{IEEEtran}
\bibliography{root.bbl}

\begin{thebibliography}{10}
\providecommand{\url}[1]{#1}
\csname url@samestyle\endcsname
\providecommand{\newblock}{\relax}
\providecommand{\bibinfo}[2]{#2}
\providecommand{\BIBentrySTDinterwordspacing}{\spaceskip=0pt\relax}
\providecommand{\BIBentryALTinterwordstretchfactor}{4}
\providecommand{\BIBentryALTinterwordspacing}{\spaceskip=\fontdimen2\font plus
\BIBentryALTinterwordstretchfactor\fontdimen3\font minus
  \fontdimen4\font\relax}
\providecommand{\BIBforeignlanguage}[2]{{%
\expandafter\ifx\csname l@#1\endcsname\relax
\typeout{** WARNING: IEEEtran.bst: No hyphenation pattern has been}%
\typeout{** loaded for the language `#1'. Using the pattern for}%
\typeout{** the default language instead.}%
\else
\language=\csname l@#1\endcsname
\fi
#2}}
\providecommand{\BIBdecl}{\relax}
\BIBdecl

\bibitem{carla}
A.~Dosovitskiy, G.~Ros, F.~Codevilla, A.~Lopez, and V.~Koltun, ``{CARLA}: {An}
  open urban driving simulator,'' in \emph{Proceedings of the 1st Annual
  Conference on Robot Learning}, 2017, pp. 1--16.

\bibitem{vtd}
\BIBentryALTinterwordspacing
``{VTD} homepage.'' 2019. [Online]. Available:
  \url{https://vires.com/vtd-vires-virtual-test-drive}
\BIBentrySTDinterwordspacing

\bibitem{ngsim}
\BIBentryALTinterwordspacing
``{NGSIM} homepage. {FHWA}.'' 2005-2006. [Online]. Available:
  \url{http://ngsim.fhwa.dot.gov.}
\BIBentrySTDinterwordspacing

\bibitem{highd}
R.~Krajewski, J.~Bock, L.~Kloeker, and L.~Eckstein, ``The highd dataset: A
  drone dataset of naturalistic vehicle trajectories on german highways for
  validation of highly automated driving systems,'' in \emph{2018 IEEE 21st
  International Conference on Intelligent Transportation Systems (ITSC)}, 2018.

\bibitem{kitti}
A.~Geiger, P.~Lenz, C.~Stiller, and R.~Urtasun, ``Vision meets robotics: The
  kitti dataset,'' \emph{The International Journal of Robotics Research},
  vol.~32, no.~11, pp. 1231--1237, 2013.

\bibitem{oxford}
\BIBentryALTinterwordspacing
W.~Maddern, G.~Pascoe, C.~Linegar, and P.~Newman, ``{1 Year, 1000km: The Oxford
  RobotCar Dataset},'' \emph{The International Journal of Robotics Research
  (IJRR)}, vol.~36, no.~1, pp. 3--15, 2017. [Online]. Available:
  \url{http://dx.doi.org/10.1177/0278364916679498}
\BIBentrySTDinterwordspacing

\bibitem{bdd}
F.~Yu, W.~Xian, Y.~Chen, F.~Liu, M.~Liao, V.~Madhavan, and T.~Darrell,
  ``Bdd100k: A diverse driving video database with scalable annotation
  tooling,'' \emph{arXiv preprint arXiv:1805.04687}, 2018.

\bibitem{driverintent}
M.~Liebner, M.~Baumann, F.~Klanner, and C.~Stiller, ``Driver intent inference
  at urban intersections using the intelligent driver model,'' in \emph{2012
  IEEE Intelligent Vehicles Symposium}.\hskip 1em plus 0.5em minus 0.4em\relax
  IEEE, 2012, pp. 1162--1167.

\bibitem{generalizableintention}
D.~J. Phillips, T.~A. Wheeler, and M.~J. Kochenderfer, ``Generalizable
  intention prediction of human drivers at intersections,'' in \emph{2017 IEEE
  Intelligent Vehicles Symposium (IV)}.\hskip 1em plus 0.5em minus 0.4em\relax
  IEEE, 2017, pp. 1665--1670.

\bibitem{ecc}
G.~D. Evangelidis and E.~Z. Psarakis, ``Parametric image alignment using
  enhanced correlation coefficient maximization,'' \emph{IEEE Transactions on
  Pattern Analysis and Machine Intelligence}, vol.~30, no.~10, pp. 1858--1865,
  2008.

\bibitem{RANSAC}
M.~A. Fischler and R.~C. Bolles, ``Random sample consensus: a paradigm for
  model fitting with applications to image analysis and automated
  cartography,'' in \emph{Readings in computer vision}.\hskip 1em plus 0.5em
  minus 0.4em\relax Elsevier, 1987, pp. 726--740.

\bibitem{retinanet}
T.-Y. Lin, P.~Goyal, R.~Girshick, K.~He, and P.~Doll{\'a}r, ``Focal loss for
  dense object detection,'' \emph{IEEE transactions on pattern analysis and
  machine intelligence}, 2018.

\bibitem{kf}
Y.~Chan, A.~Hu, and J.~Plant, ``A kalman filter based tracking scheme with
  input estimation,'' \emph{IEEE transactions on Aerospace and Electronic
  Systems}, no.~2, pp. 237--244, 1979.

\bibitem{ECC_Downsample}
S.~Overflow, ``cv2 motion euclidean for the warpmode in ecc image alignment
  method.''

\bibitem{coco}
H.~Caesar, J.~Uijlings, and V.~Ferrari, ``Coco-stuff: Thing and stuff classes
  in context,'' in \emph{Computer vision and pattern recognition (CVPR), 2018
  IEEE conference on}.\hskip 1em plus 0.5em minus 0.4em\relax IEEE, 2018.

\bibitem{ronneberger2015u}
O.~Ronneberger, P.~Fischer, and T.~Brox, ``U-net: Convolutional networks for
  biomedical image segmentation,'' in \emph{International Conference on Medical
  image computing and computer-assisted intervention}.\hskip 1em plus 0.5em
  minus 0.4em\relax Springer, 2015, pp. 234--241.

\bibitem{iglovikov2018ternausnet}
V.~Iglovikov and A.~Shvets, ``Ternausnet: U-net with vgg11 encoder pre-trained
  on imagenet for image segmentation,'' \emph{arXiv preprint arXiv:1801.05746},
  2018.

\bibitem{imagenet}
O.~Russakovsky, J.~Deng, H.~Su, J.~Krause, S.~Satheesh, S.~Ma, Z.~Huang,
  A.~Karpathy, A.~Khosla, M.~Bernstein, A.~C. Berg, and L.~Fei-Fei, ``{ImageNet
  Large Scale Visual Recognition Challenge},'' \emph{International Journal of
  Computer Vision (IJCV)}, vol. 115, no.~3, pp. 211--252, 2015.

\end{thebibliography}

\end{document}